# Normalization based K means Clustering Algorithm

Deepali Virmani[1], Shweta Taneja[2], Geetika Malhotra[3]

[1]Department of Computer Science,Bhagwan Parshuram Institute of Technology,New Delhi
Email:deepalivirmani@gmail.com

[2]Department of Computer Science,Bhagwan Parshuram Institute of Technology,New Delhi
Email:shweta_taneja08@yahoo.co.in

[3]Department of Computer Science,Bhagwan Parshuram Institute of Technology,New Delhi
Email:geets002@gmail.com

***Abstract-*** *K-means is an effective clustering technique used to separate similar data into groups based on initial centroids of clusters. In this paper, Normalization based K-means clustering algorithm(N-K means) is proposed. Proposed N-K means clustering algorithm applies normalization prior to clustering on the available data as well as the proposed approach calculates initial centroids based on weights. Experimental results prove the betterment of proposed N-K means clustering algorithm over existing K-means clustering algorithm in terms of complexity and overall performance.*

***Keywords-*** *Clustering, Data mining, K means, Normalization, Weighted Average*

## I.    INTRODUCTION

Data mining[7][11]or knowledge discovery is a process of analysing large amounts of data and extracting useful information. It is an important technology which is used by industries as a novel approach to mine data. Data mining tools and techniques are used to generate effective results which was earlier difficult and time consuming.Data mining is widely used in various areas like financial data analysis, retail and telecommunication industry, biological data analysis, fraud detection, spatial data analysis and other scientific applications.

Clustering is a technique of data mining in which similar objects are grouped into clusters. Clustering techniques are widely used in various domains like information retrieval, image processing,etc[1][2].There are two types of approaches in clustering: hierarchical and partitioning. In hierarchical clustering, the clusters are combined based on their proximity or how close they are. This combination is prevented when further process leads to undesirable clusters. In partition clustering approach, one dataset is separated into definite number of small sets in a single iteration[10]. The accuracy and quality of clustering results depends how the algorithms are implemented and their ability to find hidden knowledge.

There are various clustering algorithms based on the nature of generated clusters and techniques. Few of them are BIRCH(Balanced iterative reducing and clustering using hierarchies),CURE(Clustering using representatives),K-means, genetic K-means, Clara, Dbscan,Clarans etc[6]. The most widely used clustering algorithm is the K- means algorithm. This algorithm is used in many practical applications.It works by selecting the initial number of clusters and initial centroids[7][13]. We have chosen K-means algorithm over other clustering algorithms as it very efficient in processing large data sets. It often terminates at a local optimum and generates tighter clusters than hierarchical clustering, especially if clusters are globular. It is a popular algorithm because of its observable speed and simplicity.But K-means has a major disadvantage that it does not work well with clusters of different size and different density. Moreover initial centroids are chosen randomly due to which clusters produced vary from one run to another. Also various datapoints exist on which K-means takes superpolynomial time[8][5].

Different researchers have put forward various methods to improve the efficiency and time of K-

means algorithm. K-means uses the concept of Euclidean distance to calculate the centroids of the clusters. This method is less effective when new data sets are added and have no effect on the measured distance between various data objects. The computational complexity of k means algorithm is also very high[1][9].Also, K-means is unable to handle noisy data and missing values.Data preprocessing techniques are often applied to the datasets to make them more clean, consistent and noise free. Normalization is used to eliminate redundant data and ensures that good quality clusters are generated which can improve the efficiency of clustering algorithms.So it becomes an essential step before clustering as Euclidean distance is very sensitive to the changes in the differences[3].

This paper is organized as follows:in Section II, a description of the literature survey is done in which we have covered the work done by various authors to improve K-means clustering algorithm. Then in Section III, our proposed N-K means algorithm is described stepwise followed by experimental results in Section IV, where a comparison is shown between traditional K-means clustering algorithm and N-K means clustering algorithm. Lastly, the conclusions are addressed in Section V.

## II. LITERATURE SURVEY

A lot of methods and techniques have been proposed over the past few years to improve the accuracy of the algorithm and there is a need to optimize it to have good results.This section discusses the various approaches proposed by researchers to find better initial centroids in k means algorithm .

Authors[3], have proposed data preprocessing techniques like cleaning and normalization to produce optimum quality clusters. In normalization the data to be analyzed is scaled to a specific range. A modified k means algorithm is proposed which provides a solution for automatic initialization of centroids and performance is enhanced using normalization. This techqnique overcomes many drawbacks of naive k means algorithm.

Some authors have proposed their methods to identify initial centroids. In the following work[1], authors proposed a novel method to find better initial centroids as well as more accurate clusters with less computational time. This method was adopted to find weighted average score of dataset by averaging the value of attribute of each data point to generate initial centroids.

In another work, Authors[8] proposed a new k means clustering method with improved initial centre. In this method, initial cluster centres are selected and the centres are used as input to the k means. The user is not required to give the number of clusters as input.

In another research as done by authors[4], a data clustering approach is proposed which works by partitioning the space into different segments and calculating the frequency of data point in each segment and the segment which shows maximum frequency of data point has maximum chances to contain the centroid of the cluster.The authors have introduced concept of threshold distance for each cluster's centroid for comparing the distance between data point and cluster's centroid and using this method, efforts to calculate the distance between data point and cluster's centroid is minimized. This algorithm effectively decreases the complexity and makes calculations easier.

## III. PROPOSED N- K MEANS ALGORITHM

K-means algorithm can generate better results after the modification of the databases. We apply the modified algorithm with calculation of initial centroids based on weighted average score of dataset. Next, we preprocess and normalize dataset before we apply the N-K means algorithm. This proposed method works in three stages.During the first stage,data preprocessing technique is adopted that transforms raw data into understandable format. During the second stage, normalization is performed to standardize the data objects into specific range. During the third stage we apply the N-K means algorithm to generate clusters.

### 3.1 DATA PRE - PROCESSING
It is a very important step and should be adopted in clustering as this method uses concepts like constant, average, minimum, maximum, standard deviation to calculates missing values in the tuples[7]. These missing values need to be avoided for accurate results. Preprocessing involves steps like data cleaning, data integration, data transformation, data reduction and data discretization.

### 3.2 NORMALIZATION
Data Mining can generate effective results if normalization is applied to the dataset. It is a process used to standardize all the attributes of the dataset and give them equal weight so that redundant or noisy objects can be eliminated and there is valid and reliable data which enhances the accuracy of the result. K-Means algorithm uses Euclidean distance that is highly prone to irregularities in the size of various features[3]. There are various data normalization methods like Min-Max, Z-Score and Decimal Scaling. The best normalization method depends on the data to be normalized. Here, we have used Min-Max normalization technique in our algorithm because our dataset is limited and has not much variability between minimum and maximum. Min-Max normalization technique performs a linear transformation on the data. In this method, we fit the data in a predefined boundary or in a predefined interval.

### 3.3 INITIAL CENTROIDS CALCULATION

We use a uniform method to find score by taking the average of the attribute of each data point which will generate initial centroids that follow the data distribution of the given set. A sorting algorithm is applied to the score of each data point and then divided into k subsets where k is the number of clusters. Finally the nearest value of mean from each subset is taken as initial centroid. In this method we have introduced a weight with each attribute, which makes the method advantageous as it can cause enhancement of any feature of the dataset by increasing the weight related to that attribute.

The algorithm is given in Fig 1:

---

**ALGORITHM 1: Steps of N-K means Algorithm**

INPUT: A dataset with d dimensions

OUTPUT: Clusters

1. Load initial data set.
2. Find the maximum and minimum values of each feature from the dataset.
3. Normalize real scalar values of datasets with maximum and minimum values using equation :  $v' = \dfrac{v - \min(e)}{\max(e) - \min(e)}$  (1) where,

    min(e) and max(e) are the minimum and the maximum values for attribute E.

4. Pass the number of clusters and generate initial centroids using algorithm 2.
5. Generate clusters.

---

Figure 1: shows the steps of N-K means algorithm

---

**ALGORITHM 2: Initialization of centroids**

1. Calculate the average score of each data point.
   1) $d_i = x_1, x_2, x_3, x_4 \ldots x_n$
   2) $d_i(avg) = (w_1 \ast x_1 + w_2 \ast x_2 + w_3 \ast x_3 + \ldots w_m \ast x_m)/m$ where x= attribute's value , m= no of attributes, w= weight to multiply to ensure fair distribution of cluster.

2. Sort the data based on average score .

3. Divide the data based on k subsets.

4. Calculate the mean value of each subset.

5. Take the nearest possible data point of the mean as the initial centroid for each data subsets.

---

Figure 2: Algorithm to calculate initial number of centroids

## IV. EXPERIMENTAL RESULTS

Our experiment was conducted on Iris data set [12] from UCI Machine Learning Repository for evaluating the performance of N-K means clustering algorithm. In this section, we represent a comparative analysis of traditional K-means clustering algorithm with N-K means algorithm. Both the algorithms are run for different values of k. From the comparisons we can make out that N-K means algorithm outperforms the traditional K-means algorithm in terms of parameters namely execution time and speed. Hence the algorithm computationally runs faster as it executes in less number of iterations and the complexity is reduced. The results are depicted in Table 1.

Table 1: Performance Comparison of N-K means Algorithm With Existing K-means

| Value of k | Algorithm | Time taken (ms) | Speed |
|---|---|---|---|
| 1 | K means | 0.078 | 5.1 |
|   | N- K means | 0.065 | 3.5 |
| 3 | K-means | 0.094 | 6.2 |
|   | N-K means | 0.081 | 4.7 |
| 5 | K-means | 0.125 | 6.6 |
|   | N-K means | 0.103 | 5.0 |
| 7 | K-means | 0.134 | 7.2 |
|   | N-K means | 0.117 | 5.7 |

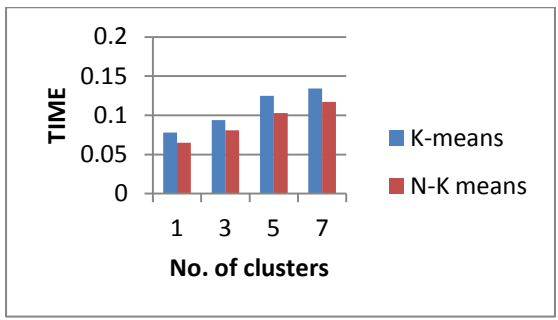

Figure 3: shows comparison between K-means and N-K means on the basis of time

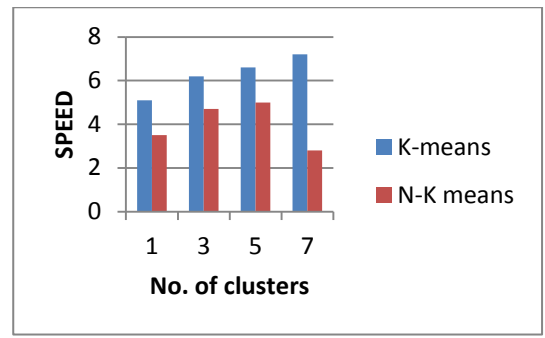

Figure 4: shows comparison between K-means and N-K means on the basis of speed

## V. CONCLUSION

The K-means clustering algorithm is widely used for clustering huge data sets. But traditional k means algorithm does not always generate good quality results as automatic initialization of centroids affects final clusters. This paper presents an efficient algorithm where we have first preprocessed our dataset based on normalization technique and then generated effective clusters. This is done by assigning weights to each attribute value to achieve standardization. Our algorithm has proved to be better than traditional K-means algorithm in terms of execution time and speed.